\documentclass{article}
\usepackage[final]{neurips_2022}
\usepackage[utf8]{inputenc} 
\usepackage[T1]{fontenc}    
\usepackage{hyperref}       
\usepackage{url}            
\usepackage{booktabs}       
\usepackage{amsfonts}       
\usepackage{nicefrac}       
\usepackage{microtype}      
\usepackage{xcolor}         
\usepackage{amsmath}
\usepackage{amssymb}
\usepackage{graphicx}
\usepackage[table]{xcolor}
\usepackage{colortbl}
\usepackage{float}

\title{Knowledge Distillation for Large Language Models}

\author{%
  Alejandro Paredes La Torre \\
  \And
  Barbara Flores \\
  \And
  Diego Rodriguez \\
  \AND
  Duke University 
}
\begin{document}

\maketitle

\begin{abstract}
We propose a resource-efficient framework for compressing large language models through knowledge distillation (KD) combined with guided chain-of-thought (CoT) reinforcement learning. Using Qwen 3B as the teacher and Qwen 0.5B as the student, we apply KD across English (Dolly-15k), Spanish (Dolly-15k), and code (BugNet and PyTorrent) datasets, with hyperparameters tuned in the English setting to optimize student performance. Across tasks, the distilled student retains a substantial portion of the teacher’s capability while remaining significantly smaller: 70–91\% in English, up to 95\% in Spanish, and up to 93.5\% Rouge-L in code. For coding tasks, integrating CoT prompting with Group Relative Policy Optimization (GRPO) using CoT-annotated Codeforces data improves reasoning coherence and solution correctness compared to KD alone. Post-training 4-bit weight quantization further reduces memory footprint and inference latency. These results show that KD combined with CoT-guided reinforcement learning can produce compact, efficient models suitable for deployment in resource-constrained settings.
\end{abstract}

\section{Introduction}
\subsection{Problem}
Training and deploying a Large Language Model (LLM) demands significant computational power, memory, and time. Large models like ChatGPT need powerful GPUs or cloud infrastructure to function effectively. This makes it challenging for small organizations, researchers with limited resources, or contexts where real-time applications are essential to use these models. To address this challenge, distillation techniques \cite{sanh2020distilbertdistilledversionbert} are applied. Specifically, Knowledge Distillation (KD) is employed, where a smaller model learns from a larger one. This approach enables achieving comparable results to the larger model while using only a fraction of the parameters, thereby improving the efficiency of model deployment. \cite{sanh2020distilbertdistilledversionbert, gu2024minillmknowledgedistillationlarge}.
\subsection{Objectives}
This study explored methods for compressing neural networks by investigating distillation techniques. Specifically, we implemented Knowledge Distillation following the approach proposed by Gu et al. (2024), to develop a more efficient version of a pre-trained LLM that requires fewer computational resources than the original large model \cite{gu2024minillmknowledgedistillationlarge}. This document test different sets of hyperparameters and data sets to assess how these changes affect the effectiveness of the model in different language domains. This helps to understand the potential for creating smaller models with comparable performance that can be used in contexts with more limited resources. As a secondary contribution, post-training quantization was applied to the distilled model to further reduce memory footprint and inference latency, facilitating on-device deployment in resource-constrained environments.

\subsection{Contributions}
Throughout the document, the following experiments were conducted: First, we conducted the Qwen Knowledge Distillation training process using the teacher model (Qwen 3B) with the student model (Qwen 0.5B) across 3 different domains, with decreasing context levels: English, Coding, and Spanish. Fine-tuning with reinforcement learning to produce valid Chain-of-thought (CoT) traces with guidance was included to perform complex tasks on coding, using Qwen Family of model capabilities.

The work here presented differs from previous works as the KD process is compared under different levels of LLMs pre-existing knowledge, using English as a benchmark to evaluate coding and Spanish tasks.
Furthermore, instead of applying CoT and RL to create foundational LLMs \cite{deepseekai2025deepseekr1incentivizingreasoningcapability}, the present work applies these techniques specifically to coding tasks. We employ the DeepSeek R1-Zero framework and leverage Guided Reinforcement Preference Optimization (GRPO) to iteratively refine model outputs \cite{deepseekai2025deepseekr1incentivizingreasoningcapability}. The training procedure was conducted on the open-r1/codeforces-cots dataset \cite{penedo2025codeforces}, which consists of CoT-annotated problem-solution pairs derived from the Codeforces competitive programming platform.
As a minor contribution, post-training quantization was applied to the distilled student model, reducing its memory footprint and inference latency to levels suitable for on-device serving in resource-constrained environments. This step demonstrates a practical pathway from distilled model to efficient edge deployment, complementing the primary knowledge distillation and reasoning objectives of this work. The code and data are available at: \url{https://github.com/AlejandroParedesLT/knowledge_distillLLM}.
\section{Related Work}

\subsection{Context}
Several studies consider similar approaches. First, the starting reference for Knowledge Distillation happened on the context of image classification, presented by Hinton et al.~\cite{hinton2015distillingknowledge}, who introduced the idea of using a large neural net model as a teacher to train a smaller student model. Another example of a distillation technique applied in image classification is the research developed by Yuan et al.~\cite{yuan2021revisitingknowledgedistillationlabel} and Zhang et al.~\cite{zhang2019your}. These papers proposed techniques that don’t require an external teacher model.~\cite{yuan2021revisitingknowledgedistillationlabel} introduced a teacher-free framework, while~\cite{zhang2019your} explored self-distillation, where a model distills knowledge from its own predictions.

In the field of natural language processing, the foundational work done by Sanh et al.~\cite{sanh2020distilbertdistilledversionbert} on their implementation of DistilBERT laid the steps to explore distillation in this field. The paper explains that by using the logits from a teacher model, in this case, BERT (Devlin et. al~\cite{devlin-etal-2019-bert}), a student model, is capable of “replicating” the embedding capabilities of the teacher model. This work differentiated from prior contributions as it was used in the pre-training phase, reducing the size of the teacher model by 40\% while retaining 97\% of its language understanding. The authors maintained the structure of the teacher model while reducing the number of layers by a factor of 2. A distillation loss over soft target probabilities of the teacher model is used as the target function to optimize. The student model was trained using the teacher model’s logits as soft labels to mimic its behavior, following a knowledge distillation approach.

Pivoting to the area of text generation, distillation has also been explored for models based on decoder-only transformers. In this research field, KD of Large Language Models by Gu et al.~\cite{gu2024minillmknowledgedistillationlarge} is a novel work. The authors use pre-trained large open-source LLM models such as GPT-2-1.5B, GPT-J 6B, and OPT 13B as teacher models, and use their pre-trained student counterparts to estimate their performance in English, successfully creating compact models with high performance. This approach was pushed further by DeepSeek integration of Chain-of-Thought (CoT) prompting, and Reinforcement Learning (RL), to guide LLMs in solving complex task with smaller models. \cite{deepseekai2025deepseekr1incentivizingreasoningcapability}.

\section{Methodology}
The Transformer architecture serves as the foundational framework for the development of state-of-the-art autoregressive generative models. At its core, this architecture leverages the scaled dot-product attention mechanism \cite{vaswani2023attentionneed}, which is defined as follows.
\begin{equation}
\text{Attention}(Q_i, K, V) = \sum_{j=1}^{n} \text{softmax}\left(\frac{Q_i K_j^\top}{\sqrt{d_k}}\right) V_j
\end{equation}
 Furthermore, the target function of decode-only models can be expressed as:
\begin{equation}
\mathcal{L}_{\text{LM}}(\theta) = - \sum_{t=1}^{T} \log p_{\theta}(x_t | x_{<t})
\end{equation}
where:
\begin{itemize}
    \item $x_{1:T}$ is the token sequence.
    \item $\theta$ are the model parameters.
    \item $p_{\theta}(x_t | x_{<t})$ is the probability assigned to token $x_t$ given its preceding context.
\end{itemize}

\subsection{Knowledge Distillation}

For our experiments, the target function in the context of LLMs can be expressed as:
\begin{equation}
\mathcal{L}_{\text{KD}} = - \sum_{t=1}^{T} \sum_{v \in \mathcal{V}} P_T(v \mid x_{<t}) \log P_S(v \mid x_{<t})
\end{equation}

where:
\begin{itemize}
  \item $\mathcal{V}$ is the vocabulary,
  \item $P_T$ is the teacher model's predicted distribution,
  \item $P_S$ is the student model's predicted distribution.
\end{itemize}
With respect to the KL Divergence Term (Soft Targets)
\begin{equation}
\text{KL}(p_T^{(\tau)} \| p_S^{(\tau)}) = \sum_{t=1}^{T} \sum_{v \in \mathcal{V}} p_T^{(\tau)}(v | x_{<t}) \log \frac{p_T^{(\tau)}(v | x_{<t})}{p_S^{(\tau)}(v | x_{<t})}
\end{equation}

where $\mathcal{V}$ is the vocabulary.

\begin{figure}[t]
    \centering
    \includegraphics[width=0.7\textwidth]{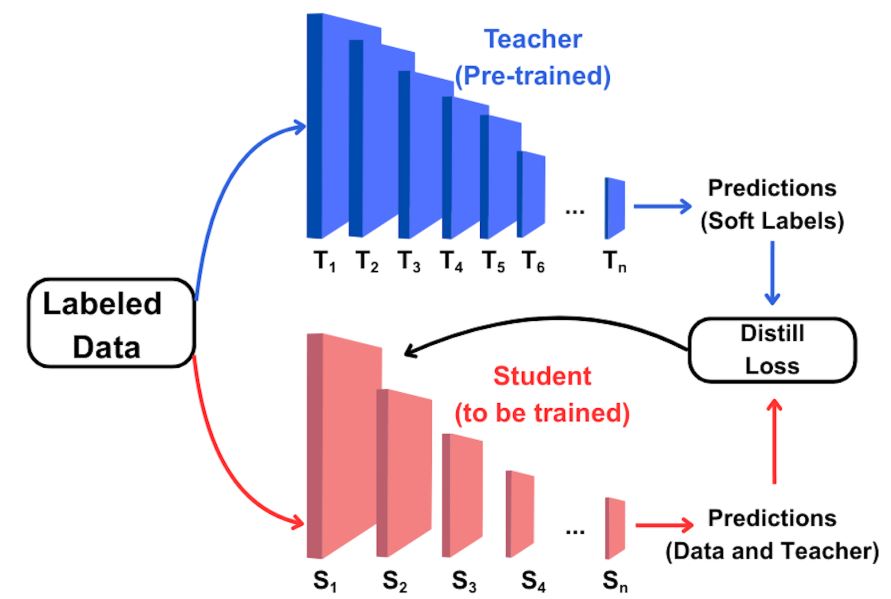}
    \caption{Distillation process with a bigger SFT model (Teacher) and a smaller model (Student) that adjusts weights on the joint distillation loss. The student model uses the sequence generated by the teacher to learn the distribution of the task. }
    \label{fig:Figure_distillation}
\end{figure}

\subsection{Chain of Thought}
In order to improve the performance of the coding generation task in this work, we use reinforcement learning with Group Relative Policy Optimization \cite{deepseekai2025deepseekr1incentivizingreasoningcapability} to force the model to generate long reasoning traces. The objective function for this task is defined as:

\begin{equation}
\begin{aligned}
J_{\text{GRPO}}(\theta) = \mathbb{E}_{q \sim P(Q), \{o_i\}_{i=1}^G \sim \pi_{\theta_{\text{old}}}(O|q)} \Bigg[
\frac{1}{G} \sum_{i=1}^G \Bigg(
\min\Big(
r_i A_i,\ 
\text{clip}(r_i,\ 1 - \epsilon,\ 1 + \epsilon) A_i
\Big)
\Bigg)
\Bigg] - \beta D_{\text{KL}}(\pi_\theta \| \pi_{\text{ref}})
\end{aligned}
\end{equation}

where the KL divergence term is computed as:

\begin{equation}
D_{\text{KL}}(\pi_\theta \| \pi_{\text{ref}}) =
\frac{\pi_{\text{ref}}(o_i | q)}{\pi_{\theta}(o_i | q)}
- \log\left(\frac{\pi_{\text{ref}}(o_i | q)}{\pi_{\theta}(o_i | q)}\right) - 1
\end{equation}

The advantage $A_i$ is computed using the group rewards $\{r_1, r_2, \ldots, r_G\}$ as:

\begin{equation}
A_i = \frac{r_i - \text{mean}(\{r_1, r_2, \ldots, r_G\})}{\text{std}(\{r_1, r_2, \ldots, r_G\})}
\end{equation}

\subsection{Post-Training Quantization}
Post-Training Quantization (PTQ) is a model compression technique applied to a pre-trained model without requiring any additional training or fine-tuning. In this work, PTQ is employed as a minor contribution to further reduce the memory footprint and inference latency of the distilled student model, enabling its deployment in resource-constrained, on-device environments.

Formally, quantization maps a full-precision floating-point weight $w \in \mathbb{R}$ to a low-bit integer representation $\hat{w} \in \mathbb{Z}$ via a uniform affine mapping:
\begin{equation}
\hat{w} = \text{clamp}\left(\left\lfloor \frac{w}{s} \right\rceil + z,\ q_{\min},\ q_{\max}\right)
\end{equation}
where $s > 0$ is the scale factor, $z$ is the zero-point offset, $\lfloor \cdot \rceil$ denotes rounding to the nearest integer, and $[q_{\min}, q_{\max}]$ defines the target integer range corresponding to the chosen bit-width. The dequantized approximation is then recovered as:
\begin{equation}
\tilde{w} = s \cdot (\hat{w} - z)
\end{equation}
The quantization error $\|w - \tilde{w}\|$ is minimized by calibrating $s$ and $z$ over a small representative dataset. In this work, 4-bit weight quantization (W4) is applied to all linear layers of the distilled model using the GPTQ framework \cite{frantar2023gptqaccurateposttrainingquantization}, which minimizes the layer-wise reconstruction error via approximate second-order information, achieving a substantial reduction in model size with negligible degradation in task performance.
\section{Experiments}
\subsection{Knowledge Distillation English}

In this experiment, the GPT-2 model with 1.5B parameters, pre-trained for text generation, was used as the teacher model. Specifically, the publicly available MiniLLM/teacher-qwen-0.5B checkpoint \cite{gu2024minillmknowledgedistillationlarge}.The student model, with 125M parameters, was trained using knowledge distillation (KD). The training was done on the Databricks Dolly 15k dataset, which is designed for question-answering tasks. The model was trained for 5 epochs, with the learning rate adjusted to analyze its impact on performance.

    \begin{figure}[H]
        \centering
        \includegraphics[width=0.95\textwidth]{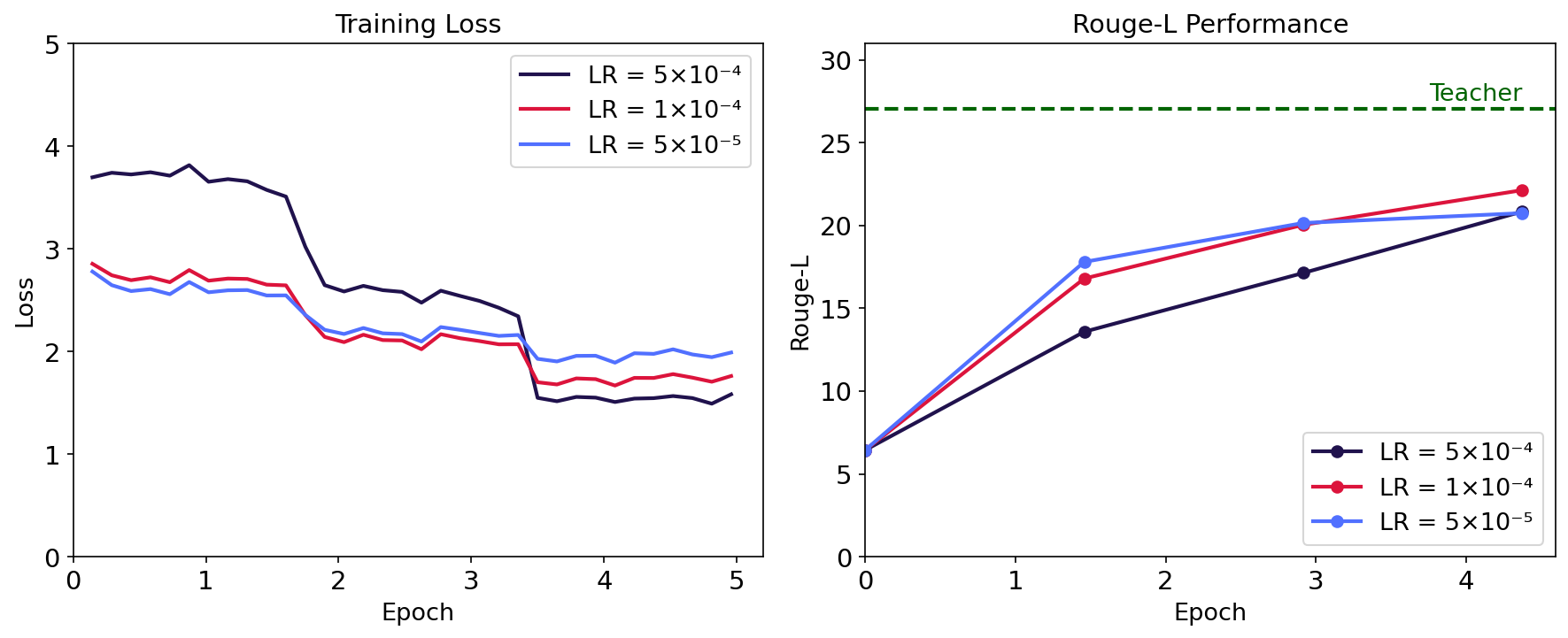}
        \caption{Performance of the Qwen0.5B model on the English Dolly dataset with varying learning rates.}
        \label{fig:kd_english} 
    \end{figure}
    
As shown in Figure \ref{fig:kd_english}, the performance of the Qwen 0.5B model with different learning rates during knowledge distillation is observed. While different learning rates exhibit varying convergence patterns, we can observe that the learning rate of $5 \times 10^{-4}$ results in the greatest loss reduction after epoch 3 with the best Rouge-L results, aligning with the LR recommended in the original paper \cite{gu2024minillmknowledgedistillationlarge}.

\subsection{Knowledge Distillation Spanish}

The Qwen 3B was fine-tuned for 10 epochs on the Spanish subset of the Dolly multilingual dataset to serve as the teacher model. The student model, was then trained using both supervised fine-tuning (SFT) and knowledge distillation (KD) approaches.

    \begin{figure}[H]
        \centering
        \includegraphics[width=.95\linewidth]{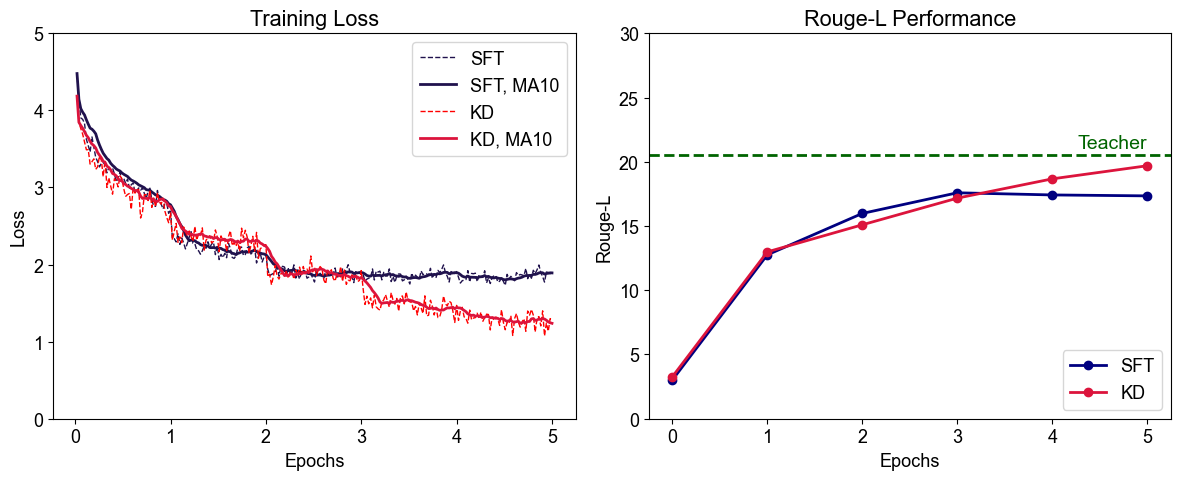}
        \caption{Performance of the Qwen 0.5B model on Spanish Dolly Multilingual dataset.}
        \label{fig:kd_spanish}
    \end{figure}

As shown in Figure~\ref{fig:kd_spanish}, the knowledge-distilled student outperformed the supervised fine-tuned counterpart, particularly in early training stages. The KD-trained model achieved both lower training loss and higher Rouge-L scores. Notably, its performance closely approached that of the teacher model with a 10\% of the total available parameters, suggesting that knowledge distillation is especially effective in scenarios with limited domain-specific knowledge by the model.

\subsection{Knowledge distillation for code correction and code generation}
A first experiment for the task of code correction using the bugnet dataset \cite{jercan2023bugdetection}. This dataset consists on examples of incorrect code, the standard console output describing the error and the corresponding correct code. A Qwen2 3B was used as the teacher model performing a supervised fine tuning and then using Qwen2 0.5B base as the student model for the knowledge distillation task. 

A second experiment on knowledge distillation focusing on code generation was performed using the same set of models as the code correction task. The dataset called pytorrent \cite{bahrami2021pytorrent} contains pairs of natural Language and programming language curated from very high quality libraries hosted in github repositories. The same steps as the code correction experiment were used for this dataset.

Extending to a third experiment, the pytorrent dataset was used to perform a knowledge distilation using the Qwen2.5 model \cite{qwen2025qwen25technicalreport}. As per the first experiment, a supervised fine tuning to the teacher model, in this case Qwen2.5 1.5B and a knowledge distillation for a student model, in this case Qwen2.5 0.5B, were performed. Leveraging the longer input span that the qwen family of models have, a reinforcement learning framework was proposed to improve the  performance of the distilled student model. All distilled student models were further quantized to reduce memory footprint and inference latency, facilitating on-device serving in resource-constrained environments.

    \begin{figure} [H]
        \centering
        \includegraphics[width=0.95\textwidth]{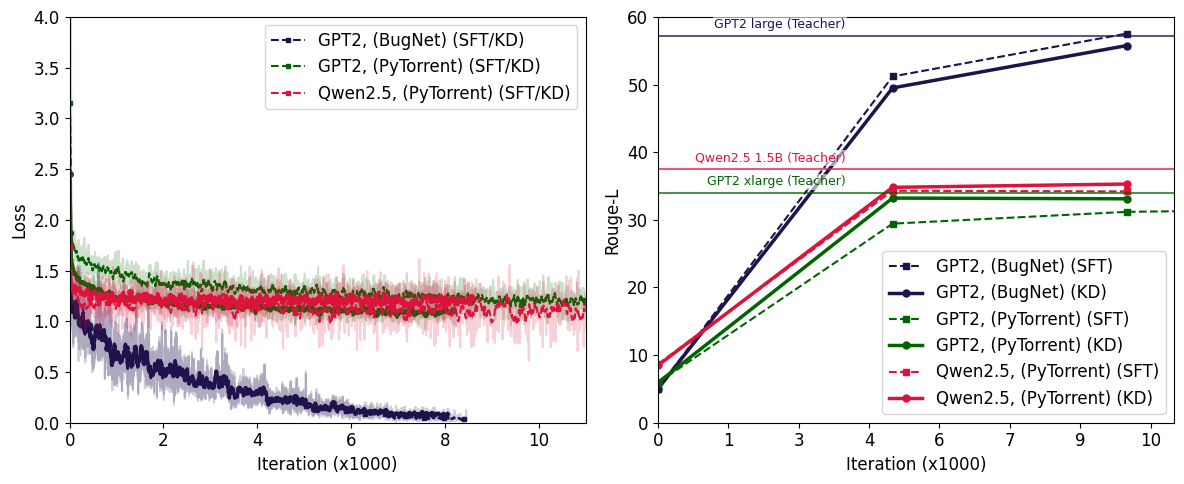}
        \caption{Performance of Qwen2 0.5B model vs a SFT baseline on the bugnet dataset, Qwen2 0.5B model vs a SFT baseline on the pytorrent dataset and Qwen2.5  model vs a SFT baseline on the pytorrent dataset}
        \label{fig:kd_experiment}
    \end{figure}

\subsection{Reinforcement Learning for Chain of Thought}

The reinforcement learning experiment was set up using a suite of reward functions tailored to competitive programming scenarios. The goal was to optimize generated solutions to adhere to correct problem-solving procedures and strict output formatting, as typically required in automated code evaluation systems. The reward signals consisted on several functions evaluating keyword consistency on <think>...</think> subsequences, other keywords such as `Step *`, overall correctness and length generation, forcing the model to be precise, think procedurally and produce meaningful long steps to reason.

    \begin{figure}[H]
        \centering
        \includegraphics[width=\textwidth]{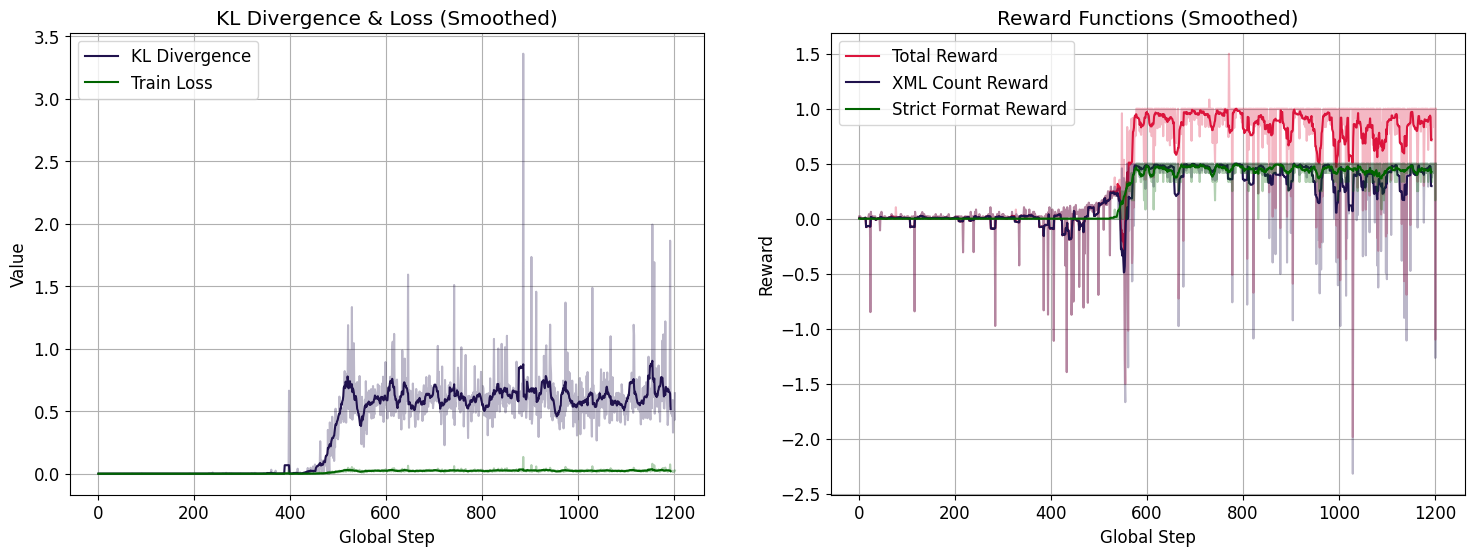}
        \caption{Training sequence of Reinforcement Learning using Chain of thought. The goal is for the model to modify it's original data distribution, short generated sequences, to longer sequences of text with specific keywords and distribution. Shown in the plot are the KL divergence which is going to increase (shift away from short generations) on the left and an increase of the reward functions.}
        \label{fig:rltraining}
    \end{figure}
\section{Conclusions}

\definecolor{pastelgreen}{RGB}{204,255,204}
\definecolor{pastelyellow}{RGB}{255,255,204}
\definecolor{pastelred}{RGB}{255,204,204}
Table \ref{tab:summary_table} summarizes the performance of the model under different experiments and models.

\begin{table}[H]
\centering
\scriptsize
\begin{tabular}{p{1.8cm} p{1.8cm}|p{1.8cm} p{1.8cm} p{1.8cm}|p{1.8cm}}
\hline
\textbf{Task} & \textbf{Prompt} & \textbf{Base} & \textbf{SFT} & \textbf{KD} & \textbf{Teacher} \\
\hline
English - Closed Question &
What is the capital of the United Kingdom? &
\cellcolor{pastelred}What is the capital of the United Kingdom? (6.6) &
\cellcolor{pastelgreen}The capital of the United Kingdom is London. (23) &
\cellcolor{pastelgreen}The capital of the United Kingdom is London. (20.4) &
\cellcolor{pastelgreen}London is the capital of the United Kingdom. (27.1) \\
\hline
Spanish - Closed Question &
¿Cuál es la capital de Reino Unido? &
\cellcolor{pastelred}Cuál has a lot to offer so you'll want to check them out. (3.1) &
\cellcolor{pastelred}El capital de Reino Unido es la Ópera, lo que no es el nombre del Reino Unido.  (17.4) &
\cellcolor{pastelyellow}La capital de Reino Unido es Nueva York. (19.7) &
\cellcolor{pastelgreen}Londres es la capital de Reino Unido.(20.6) \\
\hline
Code (Qwen) pytorrent &
Write a for loop from 1 to 10 &
\cellcolor{pastelred}The output is then shown in the output of the second loop. (5.9) &
\cellcolor{pastelred}df = Statistics. stats\_interval('interval', 'minutes')
    p = p.std() (31.2) &
\cellcolor{pastelyellow}for j in range( np.ceil(1 - np.sqrt( 1)), 2:-np.ceil(0) + (np.pi / 2):(33.1) &
\cellcolor{pastelgreen}for i in range(1, 10):
        if num >= 1:
logger.error('
(32.7) \\
\hline
Code (Qwen2.5) pytorrent &
DFS in Python &
\cellcolor{pastelred} - list all visited nodes using 'depth\_first\_search (8.7) &
\cellcolor{pastelyellow}def search(self, start=None, path=[], visit\_func = None,
            visit\_args=(),(34.4) &
\cellcolor{pastelyellow} def \_next(x, d): x = next(x) return [x] + [y for y in next(x)[:d]]
(35.5) &
\cellcolor{pastelgreen}def DFS(D= Graph ( )):      ...     D.add\_edge(5,0)
      ...     n\_d=[0]
      ...     while n\_d:
(37.5) \\
\hline
\end{tabular}
\caption{Rouge-L in parenthesis. Green: correct, Yellow: close, Red: incorrect.}
\label{tab:summary_table}
\end{table}

The performance achieved varies greatly depending on the implicit knowledge already available on the large language model as well as the technique employed. On the case of full knowledge the model achieve better performance faster than under partial knowledge demonstrated on the training under the different language dataset. This is further confirmed by the coding experiment, where the model was able to achieve even a more complex task when CoT was added into the training process. 

Overall, we show that knowledge distillation can compress large LLMs into lightweight students without sacrificing core performance, and that integrating CoT prompting with reinforcement learning unlocks substantially richer reasoning—even on specialized task, like coding.

\bibliographystyle{plainnat}
\bibliography{references}
\section{Appendix}

\subsection*{Token-wise Knowledge Distillation for Autoregressive Language Models}

Let $x_{1:T} = (x_1, \dots, x_T)$ denote a sequence of tokens and let $V$ be the vocabulary. An autoregressive language model parameterized by $\theta$ factorizes the sequence probability as
\[
p_\theta(x_{1:T}) = \prod_{t=1}^{T} p_\theta(x_t \mid x_{<t})\\
\]
\text{where $x_{<t}$ denotes the prefix $(x_1, \dots, x_{t-1})$.}

In knowledge distillation, a pretrained teacher model with distribution $p_T(\cdot \mid x_{<t})$ guides a student model $p_S(\cdot \mid x_{<t})$. At each timestep $t$, both teacher and student produce a probability distribution over $V$. The token-wise distillation objective is defined as the Kullback–Leibler divergence between teacher and student distributions:

\[
\mathcal{L}*{\mathrm{KD}} = \sum*{t=1}^{T} D_{\mathrm{KL}}\big(p_T(\cdot \mid x_{<t}) ,|, p_S(\cdot \mid x_{<t})\big).
\]

Expanding the KL divergence yields

\[
D_{\mathrm{KL}}(p_T ,|, p_S)
\sum_{v \in V}
p_T(v \mid x_{<t})
\log
\frac{p_T(v \mid x_{<t})}{p_S(v \mid x_{<t})}.
\]

Therefore,

\[
\mathcal{L}_{\mathrm{KD}}
\sum_{t=1}^{T}
\sum_{v \in V}
p_T(v \mid x_{<t})
\log p_T(v \mid x_{<t})
\sum_{t=1}^{T}
\sum_{v \in V}
p_T(v \mid x_{<t})
\log p_S(v \mid x_{<t}).
\]

Since the entropy term of the teacher distribution does not depend on student parameters, minimizing the KL divergence is equivalent to minimizing the token-level cross-entropy with soft targets:

\[
\mathcal{L}_{\mathrm{KD}}
*
\sum_{t=1}^{T}
\sum_{v \in V}
p_T(v \mid x_{<t})
\log p_S(v \mid x_{<t}).
\]

This formulation follows the original distillation framework of Hinton et al. \cite{hinton2015distillingknowledge}, where the student is trained to match the full softmax distribution of the teacher rather than only the ground-truth label. The soft targets encode dark knowledge in the relative probabilities assigned to non-maximum tokens, which improves generalization.

\subsection*{MiniLLM Framework}

MiniLLM proposes an alternative formulation of knowledge distillation for large language models based on reverse Kullback–Leibler divergence \cite{gu2024minillmknowledgedistillationlarge}. Instead of minimizing the forward divergence
\[
D_{\mathrm{KL}}(p_T ,|, p_S),
\]
MiniLLM minimizes the reverse divergence
\[
D_{\mathrm{KL}}(p_S ,|, p_T).
\]

At the sequence level, let $q_\theta(o \mid q)$ denote the student distribution over complete outputs $o$ given prompt $q$, and let $p_T(o \mid q)$ denote the teacher distribution. The objective is

\[
\mathcal{L}_{\mathrm{MiniLLM}}
D_{\mathrm{KL}}\big(q_\theta(\cdot \mid q) ,|, p_T(\cdot \mid q)\big)
\mathbb{E}*{o \sim q*\theta}
\left[
\log \frac{q_\theta(o \mid q)}{p_T(o \mid q)}
\right].
\]

This objective is optimized using on-policy sampling from the student. For each prompt, the student generates outputs according to its current policy, and the teacher provides log-probability evaluations of those outputs. The gradient of the reverse KL can be written as a policy-gradient estimator:

\[
\nabla_\theta \mathcal{L}_{\mathrm{MiniLLM}}
\mathbb{E}*{o \sim q*\theta}
\left[
\nabla_\theta \log q_\theta(o \mid q)
\left(
\log q_\theta(o \mid q)
\log p_T(o \mid q)
\right)
\right].
\]

This formulation treats distillation as a reinforcement learning problem where the reward signal is derived from the teacher likelihood. The reverse KL is mode-seeking and encourages the student to concentrate probability mass on high-probability teacher outputs. Empirically, MiniLLM demonstrates improved generation fidelity compared to conventional token-level forward KL distillation \cite{gu2024minillmknowledgedistillationlarge}.

\subsection*{Group Relative Policy Optimization (GRPO) for Chain-of-Thought}

Group Relative Policy Optimization is introduced in the DeepSeek-R1 framework for reasoning-focused reinforcement learning \cite{deepseekai2025deepseekr1incentivizingreasoningcapability}. Let $\pi_\theta(o \mid q)$ denote the policy model. For each prompt $q$, a group of $G$ outputs ${o_i}*{i=1}^{G}$ is sampled from the old policy $\pi*{\theta_{\mathrm{old}}}$.

Each output receives a scalar reward $r_i$. The rewards are normalized within the group to compute advantages:

\[
A_i
\frac{
r_i - \frac{1}{G} \sum_{j=1}^{G} r_j
}{
\mathrm{std}(r_1, \dots, r_G)
}.
\]

Let

\[
\rho_i
\frac{
\pi_\theta(o_i \mid q)
}{
\pi_{\theta_{\mathrm{old}}}(o_i \mid q)
}.
\]

The GRPO objective to be maximized is

\[
J_{\mathrm{GRPO}}(\theta)
\mathbb{E}
\left[
\frac{1}{G}
\sum_{i=1}^{G}
\min
\left(
\rho_i A_i,
\mathrm{clip}(\rho_i, 1-\epsilon, 1+\epsilon) A_i
\right)
\right]
\beta
D_{\mathrm{KL}}
\big(
\pi_\theta(\cdot \mid q)
,|,
\pi_{\mathrm{ref}}(\cdot \mid q)
\big),
\]

where $\epsilon$ is the clipping parameter and $\beta$ controls the strength of the KL regularization toward a reference policy $\pi_{\mathrm{ref}}$.

The KL term is defined as

\[
D_{\mathrm{KL}}
\big(
\pi_\theta ,|, \pi_{\mathrm{ref}}
\big)
\mathbb{E}*{o \sim \pi*\theta}
\left[
\log
\frac{
\pi_\theta(o \mid q)
}{
\pi_{\mathrm{ref}}(o \mid q)
}
\right].
\]

GRPO differs from standard PPO by replacing a learned value function baseline with group-relative normalization of rewards. The zero-mean property of the normalized advantages stabilizes training and removes the need for an explicit critic network. In the context of chain-of-thought training, rewards are assigned based on correctness and reasoning quality, and the objective shifts the policy toward higher-reward reasoning trajectories while controlling deviation from the reference model \cite{deepseekai2025deepseekr1incentivizingreasoningcapability}.
\end{document}